\documentclass{article}

\usepackage{PRIMEarxiv}

\usepackage[utf8]{inputenc} 
\usepackage[T1]{fontenc}    
\usepackage{hyperref}       
\usepackage{url}            
\usepackage{booktabs}       
\usepackage{amsfonts}       
\usepackage{nicefrac}       
\usepackage{microtype}      
\usepackage{lipsum}
\usepackage{fancyhdr}       
\usepackage{graphicx}       
\graphicspath{{media/}}     
\usepackage{multirow}
\usepackage[normalem]{ulem}
\useunder{\uline}{\ul}{}
\usepackage{arydshln}

\title{A Survey of Spanish Clinical Language Models
}

\author{
  Guillem García Subies \\
  Universidad Carlos III de Madrid \\
  Instituto de Ingeniería del Conocimiento \\
  \texttt{100500844@alumnos.uc3m.es, guillem.garcia@iic.uam.es} \\
   \And
  Álvaro Barbero Jiménez \\
  Instituto de Ingeniería del Conocimiento \\
  \texttt{alvaro.barbero@iic.uam.es} \\
   \And
  Paloma Martínez Fernández\\
  Universidad Carlos III de Madrid\\
  \texttt{pmf@inf.uc3m.es} \\
}

\begin{document}
\maketitle

\begin{abstract}
This survey focuses in encoder Language Models for solving tasks in the clinical domain in the Spanish language. We review the contributions of 17 corpora focused mainly in clinical tasks, then list the most relevant Spanish Language Models and Spanish Clinical Language models. We perform a thorough comparison of these models by benchmarking them over a curated subset of the available corpora, in order to find the best-performing ones; in total more than 3000 models were fine-tuned for this study. All the tested corpora and the best models are made publically available in an accessible way, so that the results can be reproduced by independent teams or challenged in the future when new Spanish Clinical Language models are created.
\end{abstract}


\section{Introduction}


The enormous wealth of data present in the Electronic Health Records (EHR) opens up multiple possibilities for its use not only in biomedical research (secondary use) but also in clinical practice (primary use). Transforming unstructured data (as clinical notes) into structured data of EHR offers researchers and healthcare professionals the possibility of working with clinical data of higher quality and accuracy because redundancy of information has been eliminated, data is validated, and also integrated in databases, thus allowing accessing using structured queries. The most frequent uses of EHR in research, in addition to coding systems for billing purposes (e.g., ICD-10 coding of diagnoses in discharge reports), range from exploring new diagnostic and therapeutic solutions, to evaluating patient outcomes, identifying populations at risk, developing databases and repositories, among many others.

Processing data coming from EHR has been one the most important challenges in natural language processing (NLP). The application of NLP techniques in analysing information contained in the unstructured fields (free text) of the EHR makes it possible to extract the relevant information (expressed in concepts, relationships and events) and its transformation into a structured format. In this way, the structured information obtained can be stored in repositories integrated within the EHR itself, which would open up the possibility for its automatic exploitation, facilitating the development of decision support tools, clinical practice guidelines, support tools for the development of epidemiological studies, among many other applications.

Apart from the inherent complexities of understanding text, clinical text has additional peculiarities. While documents published in scientific journals and books have undergone careful editing, clinical narrative texts are usually hastily written in a telegraphic style, with numerous ellipsis and spelling errors and with incorrect syntax in many cases, abuse of negations, jargon, abbreviations among others. For instance, incomplete sentences (predominance of phrases instead of complete sentences), new terms arising every day, others due to misspellings or even new terms that medical professionals invent to generate their notes. From the semantic point of view, there are words with uses in medical language that differ from the language of general use. From the contextual point of view, the abuse of ellipsis makes interpretation difficult (e.g., the paragraph: "Complaints of chest pain, increased frequency, especially with exertion. Usually associated with short respiration and nausea and vomiting" represents the entity "chest pain due to chest angina").

\subsection{Spanish language in NLP}

With $559$ million speakers, Spanish is the fourth most spoken language in the world \cite{ethnologue}. However, and despite its relevance, there are few relevant NLP resources in Spanish. For instance, at the date of the writing of this article there are $16520$ English models and $3466$ corpora in the Hugging Face Hub, but only $137$9 and $351$ respectively for Spanish \cite{hf_datasets} (a lot of those being multilingual models). This means that there are $11.35$ English models and 2.38 corpora for every million English speakers while there are only $2.30$ and $0.59$ respectively for Spanish. Another example can be the French language, with 309 million speakers, which has $4.64$ and $1.24$ French corpora for every million speakers.

Compared to English, Spanish is a highly inflectional language with a richer morphology; morphemes signify many syntactic, semantic, and grammatical aspects of words (such as gender, number, etc.). From a syntactic perspective, Spanish texts feature more subordinate clauses and lengthy sentences with a high degree of word order flexibility; for example, the subject is not restricted to appearing solely before the verb in a sentence. Spanish clinical texts have particularities that are explained below.

There is clearly a lack of resources for non-English language models, which has become even more pressing since the revolution brought about by BERT \cite{devlin2019bert} that has culminated with ChatGPT and GPT-4 \cite{gpt3, gpt4, gpt4survey}. This lack is also quantified with the big amount of biomedical and clinical language models in other languages. Just to name a few, in English we can find BioBERT \cite{Lee2019}, PubMedBERT \cite{pubmedbert}, BioGPT \cite{Luo2022} or ClinicalGPT \cite{wang2023clinicalgpt} and DrBERT \cite{labrak2023drbert} or  CamemBERT-bio \cite{CamemBERTbio} for French.

Therefore, a significant amount of research efforts are still required in the development of corpora and language models for the Spanish language, even more in the clinical domain \cite{ignat2023phd}.


\subsubsection{Spanish language in the Clinical domain}

Compared to the general Spanish language, in the Clinical domain there are more anglicisations as a result of the translation of English words used in biomedicine. Some of them are freely altered, while others are perfect replicas of the originals; for example, "adipocytokine" is also rendered as "adipocitocina", "adipocitoquina", and "adipocitokina". Additionally, the Spanish language uses accent marks that do not exist in English language, and whether or not these accent marks are preferred results in lexical variants. For example, the word "ebola" may become "ébola" or "ebola."

When translated, adjectives ending in "-al" sometimes retain their original form and other times adopt Spanish morphological conventions. For instance, "retinal" may become "retinal" or "retiniano". Greco-latin prefixes exhibit variations such as "psi-" ("psiquiatra" vs. "siquiatra"). 
In contrast to Spanish, where there are many variations, English uses hyphens between words more consistently. For example, "beta-alanine" is converted into "beta-alanina" or "beta alanina" or "alanina beta". The names of active ingredients are sometimes kept the same as they are in English and other times they are changed. For example, the name "acetaminophen" is changed to "acetaminofeno," "acetaminofén". Some terminology that refer to gender (male/female) are ambiguous ("el enzima"/"la enzima") or both are acceptable, for example, "el tiroides"(male)/ "la tiroides" (female) for the hormone "thyroid".

Abbreviations are frequently used in clinical notes, and typically Spanish and English abbreviations coexist. The term "PSA" stands for "prostate-specific antigen," which is preferable to the term "APE" (antígeno prostático específico). However, both languages use polysemic abbreviations frequently. Sentences in the two languages are syntactically highly similar (short sentences or phrases, abuse of negation particles and unconventional abbreviations, misspellings, speculation, and grammatical sentence anomalies, among other things). In conclusion, due to term replication or partial adaptation, there are more lexical variants of medical terms in Spanish than in English. Because of these factors, studying these texts requires more resources and normalization tools.

\subsection{Purpuse of the study}

The main objective of this study is to gather all the resources available to deal with clinical textual data in Spanish and benchmark the best models in order to obtain a leaderboard. Thus, we fine-tune more than $3000$ combinations for the best available models on $12$ public corpora to achieve our goal.

In the rest of this paper we present a brief summary on the evolution of NLP in the latest years in Section \ref{sec:previous}. In section \ref{sec:data} we make an overview of some of the latest corpora released in the clinical domain for the Spanish language. Then, in Section \ref{sec:models}, we list the best Spanish language models (both general and domain specific). In Section \ref{sec:meth} we explain what corpora, models and metrics we used for the benchmark. Finally, in Section \ref{sec:bench}, we present a detailed benchmark of a selection of the models and corpora in order to get insights of the situation of this research field.

\textbf{Contributions} We make the following contributions:

\begin{itemize}
    \item We create the \textbf{first public benchmark for clinical Spanish language models}.
    \item We prove that there is a lack of good quality clinical Spanish language models.
    \item We make a survey with the most relevant corpora and models released in the recent years.
\end{itemize}

\section{Previous work} \label{sec:previous}

This last decade has experienced an exponential growth in NLP. We went from word embedding models like Word2vec \cite{mikolov2013efficient} and GloVe \cite{pennington2014glove}, to gigantic language models with billions of parameters like PaLM \cite{chowdhery2022palm}.

This growth can be explained with the, also exponential, development of specific hardware dedicated to speed up the matrix calculations made in neural networks like NVIDIA's H100 GPUs \cite{nvidiatensorcore}, Google's TPUs \cite{jouppi2017indatacenter} or AWS's Trainium \cite{awstrainium}. With such an evolution in hardware, the training of neural architectures for the processing of sequences such as Recurrent Neural Networks (RNNs) \cite{hopfield1982neural, Rumelhart1986}, bidirectional RRNs \cite{bidirectional} and Long Short-term Memory (LSTM) \cite{HochSchm97} (and modifications like GRUs \cite{cho2014learning}) became practical.

These networks were used along word embeddings in order to train models for specific downstream tasks such as translation, token classification (or NER), text generation, text classification, etc. Every vector model encoded just a word and did not take into account context. Then, some models started to improve the encoding with subword information like FastText \cite{bojanowski2017enriching}. The biggest breakthrough here came in 2017 with the Transformer architecture and the contextualized embeddings \cite{vaswani2023attention}. After that a lot of new architectures appeared such as ELMo \cite{peters2018deep} and FLAIR \cite{akbik-etal-2019-flair}.

However, the one that definitely changed everything after the Transformer was BERT \cite{devlin2019bert} in 2018, a bidirectional encoder model based in the transformer that obtained much better results than the state of the art by that time. After that, hundreds of BERT-inspired models (like RoBERTa \cite{roberta} or DeBERTa \cite{he2021deberta}, just to name a few) were published \cite{zhou2023comprehensive}. Huggingface's Transformers library \cite{wolf2020huggingfaces} and subsequently their Datasets library \cite{lhoest2021datasets} and Hugginface Hub \cite{hub} helped to democratize these new architectures.

In parallel other models based in the Transformer were being created like GPT \cite{gpt} which was, opposite to BERT, just the generative part of the Transformers. After that, the architecture of GPT-2 \cite{gpt2} stacked more and more transformers layers one on the top of another. GPT-3 \cite{gpt3} proved a really big change also, although the main difference with the architecture of their predecessors was just the size. However, it obtained outstanding results even in few-show scenarios without the need of fine-tuning in demanding benchmarks like WinoGrande \cite{sakaguchi2019winogrande}. This set a precedent, after all the decade of Open Sourced models, GPT-3 would not be made available to the public, companies understood their real potential.

After GPT-3, in 2022 ChatGPT was trained using Reinforcement Learning from Human Feedback (RLHF) \cite{ouyang2022training} and suddenly, the chatbot revolution began. Companies announced their models but did not release them like GPT-4 \cite{gpt4} or LLaMA \cite{touvron2023llama}.

In our specific context of domain specific clinical models, they tried to follow the main enhancements of this ever-growing trend. For the clinical domain in Spanish language, most of the improvements were made in shared tasks such as IberLEF \cite{gonzalo2021iberlef, gonzalo2022overview} and BioNLP open shared tasks, \cite{9174741}, \cite{rivera2021analyzing}.

\section{Corpora} \label{sec:data}

NLP requires a great amount of data in order to obtain good results \cite{hoffmann2022training}, so it its very important to gather as much data as possible. When talking about NLP data we can classify it into two big groups depending whether the data is structured (labeled) or not.

\subsection{Unannotated corpora}

Corpora are very valuable in NLP because they are often very big and can be used to pre-train large language models without the need of labels. Although most of the times these corpora can be crawled online \cite{oscar}, this is not always possible in the clinical domain because data such as Electronic Health Records are not publicly available and contain personal data.

Note that this is not the case with biomedical data where PubMed papers and public clinical trials can be used \cite{carrino2021spanish}.

In this section we will review the most relevant corpora in Spanish for the clinical domain.

\subsubsection{SPACCC}

The Spanish Clinical Case Corpus (SPACCC) \cite{spaccc} is a collection of $1000$ clinical cases from SciELO. A clinician classified the original dump of clinical cases into realistic clinical cases and not suitable for this specific corpus.

The corpus has $382470$ tokens, a \texttt{cc-by-4.0} licence and is publicly available \footnote{https://zenodo.org/record/2560316}.

\subsubsection{European Clinical Case Corpus}

European Clinical Case Corpus (E3C) \cite{e3c} is a multilingual collection of clinical cases. The corpus contains clinical cases in English, French, Basque and Spanish.

We will focus in the Spanish part, which is a collection of other well known corpora such as SPACCC, NUBes (detailed down in Section \ref{data:nubes}), IULA+ (detailed down in Section \ref{data:iula}) and a small amount of newly crawled SciELO clinical cases.

Although the corpus is mainly unlabeled, it has some labeled sub-sets. Specifically, the corpus has 3 different parts:
\begin{itemize}
    \item Gold Standard: Named entities of time and factuality in THYME standard and medical entities in SNOMED-CT and ICD-10 standards. Annotated manually. This corpus has $20k$ tokens.
    \item Silver Standard: The same as above but with a bigger size and annotated automatically. This corpus has $70k$ tokens.
    \item Unnanotated data: All the data collected, about $1M$ tokens.
\end{itemize}

The corpus has a \texttt{cc-by-nc-4.0} licence and is publicly available \footnote{https://live.european-language-grid.eu/catalogue/corpus/7618}.

\begin{table}[!ht]
\centering
\begin{tabular}{llcll}
\hline
\textbf{Corpus} & \multicolumn{1}{c}{\textbf{Tokens}} &  \\ \hline
E3C                  & 1M  \\
SPACCC               & 0.38M  \\ \hline
\end{tabular}
\caption{Summary of Spanish Clinical Corpora}
\label{tab:corpora}
\end{table}

\subsection{Labeled data}

Labeled corpora are used to train and evaluate the models in downstream tasks such as text classification or NER.

Given the nature of the clinical data, there are few corpora and most of them have a relatively small size compared to corpora in other domains. Also, almost all of these corpora come from shared tasks.

A summary of all the corpora is presented in Table \ref{tab:datasets}.

\subsubsection{BARR2}

Biomedical Abbreviation Recognition and Resolution 2nd Edition (BARR2) \cite{barr2} is a NER corpus of clinical cases from SciELO where the entities are abbreviations (Short Form, SF) and their respective definition (Long Form, LF).

The corpus consists of two sub-tasks:

\begin{itemize}
    \item Detection of pairs SF/LF mentions in text. This is some kind of multi label classification given that there is no need to classify the tokens individually.
    \item Detection of SF mentions and their characters offset (NER) and generating their corresponding  LF.
\end{itemize}

While this corpus has been annotated by different experts, there are no annotation guidelines nor consistency analysis so it may not be as curated as a proper Gold Standard corpus.

The corpus has a size of $318$, $146$ and $220$ train, validation and test samples respectively. The test split comes from a bigger background unnanotated split of $2879$ unnanotated samples. The corpus has a \texttt{cc-by-4.0} licence and is publicly available \footnote{https://temu.bsc.es/BARR2/datasets.html}.

\subsubsection{CANTEMIST} \label{data:cantemist}

CANTEMIST (CANcer TExt Mining Shared Task – tumor named entity recognition) \cite{cantemist} corpus is a Named Entity Recognition (NER) corpus that focuses in tumor morphology concepts in clinical texts written in Spanish. Specifically. CANTEMIST is focused in detecting mentions of tumor morphology terms and link them to their corresponding eCIE-O-3.1 codes (the Spanish branch of International Classification of Diseases for Oncology). The corpus includes diverse clinical cases covering different cancer types, both common and rare.

The corpus provides a manually annotated Gold Standard generated by experts. There are also annotation guidelines and consistency analysis to ensure the quality of the corpus.

The corpus consists of three independent sub-tasks:

\begin{itemize}
    \item CANTEMIST-NER track: In this track, the goal is to find tumor morphology mentions in medical documents.
    \item CANTEMIST-NORM track: This track focuses on clinical concept normalization or named entity normalization. The entities are tumor morphology entity mentions along with their corresponding eCIE-O-3.1 codes.
    \item CANTEMIST-CODING track: The objective of this track is to return a ranked list of ICD-O-3 codes for each document.
\end{itemize}

The corpus has a size of $501$, $500$ and $300$ train, validation and test samples respectively. The test split comes from a bigger background unnanotated split of $4932$ unnanotated samples. The corpus has a \texttt{cc-by-4.0} licence and is publicly available \footnote{https://zenodo.org/record/3978041}.

\subsubsection{CARES} \label{data:cares}

CARES (Corpus of Anonymised Radiological Evidences in Spanish) \cite{cares} is a corpus composed of different radiology reports from several areas. These reports are annotated to hierarchically classify their ICD-10 codes. Specifically, each report has an annotation for the areas of the body that the report is talking about, their corresponding IDC-10 chapters, the ICD-10 codes and the sub-codes.

While this corpus has been annotated by different experts, there are no annotation guidelines nor consistency analysis so it may not be as curated as a proper Gold Standard corpus.

The corpus has a size of $2250$, and $966$ train and test samples respectively. The corpus has a \texttt{afl-3.0} licence and is publically available \footnote{https://huggingface.co/datasets/chizhikchi/CARES}.

\subsubsection{Chilean Waiting List Corpus}

The Chilean Waiting List Corpus (CWLC) \cite{chilean} is a NER clinical corpus made up of anonymized referrals from waiting lists in Chilean public hospitals. 48\% of the entities in the corpus are nested and the main entities are "Finding", "Procedure", "Family member", "Disease", "Body Part", "Medication" and "Abbreviation".

The corpus provides a manually annotated Gold Standard generated by experts. There are also annotation guidelines and consistency analysis to ensure the quality of the corpus.

The corpus has a size of $9000$ train samples. The corpus has a \texttt{cc-by-nc-sa-4.0} licence and is publically available \footnote{https://zenodo.org/record/7555181}.

\subsubsection{CodiEsp} \label{data:codiesp}

CodiEsp (Clinical Case Coding in Spanish Shared Task) \cite{miranda2020codiesp} is a corpus of clinical cases from a variety of medical topics, including oncology, urology, cardiology, pneumology or infectious diseases. The corpus has 3 different tasks:

\begin{itemize}
    \item CodiEsp-D: Multi-label classification of ICD10-CM (CIE10 Diagnóstico in Spanish).
    \item CodiEsp-P: Multi-label classification of ICD10-CM (CIE10 Procedimiento in Spanish).
    \item CodiEsp-X: NER task with the purpose of marking those segments in the text where evidences for the CIE10 labels can be found.
\end{itemize}

The corpus provides a manually annotated Gold Standard generated by a practicing physician and a clinical documentalist. There are also annotation guidelines and consistency analysis to ensure the quality of the corpus.

The corpus has a size of $500$, $250$ and $250$ train, development and test samples respectively. The test split comes from a bigger background unnanotated split of $2751$ unnanotated samples. The corpus has a \texttt{cc-by-4.0} licence and is publically available \footnote{https://zenodo.org/record/3837305}.

\subsubsection{CT-EBM-SP} \label{data:ctebmsp}

CT-EBM-SP (Clinical Trials for Evidence-based Medicine in Spanish) \cite{ctebmsp} is a NER corpus made up of PubMed and SciELO abstracts and clinical trials announcements published in the European Clinical Trials Register and Repositorio Español de Estudios Clínicos. Although these are not strictly clinical sources, abstracts are rather general and simple descriptions and, given the size of the corpus, it can be very useful to enhance the generalization of the models.

The corpus provides a manually annotated Gold Standard generated by experts. There are also annotation guidelines and consistency analysis to ensure the quality of the corpus.

The corpus has a size of $12600$, $4510$ and $4550$ train, development and test samples respectively. The corpus has a \texttt{cc-by-4.0} licence and is publically available \footnote{https://huggingface.co/datasets/lcampillos/ctebmsp}.

\subsubsection{DisTEMIST} \label{data:distemist}

The DisTEMIST (Disease Text Mining Shared Task) corpus \cite{distemist} is a collection of clinical cases written in Spanish, covering various medical specialties. The corpus is annotated with disease mentions (i.e. NER corpus) that have been standardized using Snomed-CT terminology. 

An expert filtered the gathered data so only relevant documents were included. After that, the data was preprocessed to delete redundant information such as figure references or citations. The corpus provides a manually annotated Gold Standard generated by experts. There are also annotation guidelines and consistency analysis to ensure the quality of the corpus.

The corpus consists of two independent sub-tasks:

\begin{itemize}
    \item DISTEMIST-entities subtrack: NER to find diseases, without classifying them into any kind of taxonomy.
    \item DISTEMIST-linking subtrack: NER to find disease mentions and assign a Snomed-CT term.
\end{itemize}

There is also a multilingual version of the corpus containing data in English, Portuguese, Catalan, Italian, French, and Romanian.

The corpus has a size of $750$ and $250$ train and test samples respectively. The test split comes from a bigger background unnanotated split of $3000$ unnanotated samples. The corpus has a \texttt{cc-by-4.0} licence and is publically available \footnote{https://zenodo.org/record/7614764}.

\subsubsection{eHealthKD} \label{data:ehealth}

eHealthKD (eHealth Knowledge Discovery) \cite{overviewehealthkd2021} is a cross-domain NER corpus with samples gathered from MedlinePlus and Wikinews. These sources create a very heterogeneous corpus that can be very useful to prove the generalization of the models. The corpus has 2 different tasks:

\begin{itemize}
    \item Subtask A: NER to recognize concepts, actions, predicates and references.
    \item Subtask B: Relation extraction from the previous NER task.
\end{itemize}

The corpus provides a manually annotated Gold Standard generated by experts. There are also annotation guidelines and consistency analysis to ensure the quality of the corpus.

The corpus has a size of $1500$, $50$ and $150$ train, development and test samples respectively. There are also $50$ and $150$ development and test samples in English, however they are not taken into account here because they are out of scope. The corpus has a \texttt{cc-by-nc-sa-4.0} licence and is publically available \footnote{https://github.com/ehealthkd/corpora/tree/master}.

\subsubsection{IULA-SCRC} \label{data:iula}

IULA Spanish Clinical Record Corpus (SCRC) \cite{marimon2017iula} is a NER corpus with negation entities extracted from various anonymized clinical records. Besides IULA-SCRC, there is also another version of the corpus annotated with NUBes' annotation guidelines called IULA+.

The corpus provides a manually annotated Gold Standard generated by experts. There are also annotation guidelines and consistency analysis to ensure the quality of the corpus.

The corpus has a size of $3194$ train samples. The corpus has a \texttt{cc-by-sa-3.0} licence and is publically available \footnote{https://github.com/Vicomtech/NUBes-negation-uncertainty-biomedical-corpus}.

\subsubsection{IxaMed-GS}

IxaMed-GS \cite{ixamedgs} is a NER corpus composed of real (anonymized) Electronic Health Records. The entities annotated are related to diseases and drugs and there are also annotations of relationships between entities indicating adverse drug reaction events.

The corpus provides a manually annotated Gold Standard generated by experts in pharmacology and pharmacovigilance. There are also annotation guidelines and consistency analysis to ensure the quality of the corpus.

The corpus has a size of $1875$, $1097$ and $995$ train, development and test samples respectively. The corpus is not publically available and it can only be obtained through an agreement with its authors.

\subsubsection{LivingNER} \label{data:livingner}

LivingNER \cite{livingner} is a corpus focused on species, pathogens and food in clinical case reports.

The corpus provides a manually annotated Gold Standard generated by experts. There are also annotation guidelines and consistency analysis to ensure the quality of the corpus.

The corpus consists of three independent sub-tasks:

\begin{itemize}
    \item subtask 1 (LivingNER – Species NER): NER to identify either a species of a disease or human entities.
    \item subtask 2 (LivingNER – Species Norm): NER to identify species of a disease and classifying them into NCBITax taxonomy.
    \item subtask 3 (LivingNER –  Clinical IMPACT): Multi-label classification. Classes: high impact (or not) and NCBI IDs.
\end{itemize}

The corpus has a size of $1000$, $500$ and $485$ train, development and test samples respectively. The test split comes from a bigger background unnanotated split of $~13000$ unnanotated samples. The corpus has a \texttt{cc-by-4.0} licence and is publically available \footnote{https://zenodo.org/record/7614764}.

\subsubsection{MEDDOCAN} \label{data:meddocan}

MEDDOCAN (Medical Document Anonymization Track) \cite{marimon2019automatic} is a corpus of clinical cases sampled from the SPACCC corpus and enriched with synthetic personal information. The NER entities of the corpus range from the personal data of a patient to even the personal data of the doctors.

The corpus provides a manually annotated Gold Standard generated by experts. There are also annotation guidelines and consistency analysis to ensure the quality of the corpus.

The corpus has a size of $500$, $250$ and $250$ train, development and test samples respectively. The test split comes from a bigger background unnanotated split of $2000$ unnanotated samples. The corpus has a \texttt{cc-by-4.0} licence and is publically available \footnote{https://huggingface.co/datasets/bigbio/meddocan}.

\subsubsection{NUBes} \label{data:nubes}

The NUBes (Negation and Uncertainty annotations in Biomedical texts in Spanish) corpus \cite{lima2020nubes} is a collection of $29,682$ sentences from anonymised documents from a Spanish private hospital annotated with negation and uncertainty. The sentences come from documents from the following kinds: Chief Complaint, Present Illness, Physical Examination, Diagnostic Tests, Surgical History, Progress Notes, and Therapeutic Recommendations.

This is the biggest negation corpus for Spanish clinical domain. Samples have been annotated for a NER task with all the negations and the uncertainty that is reflected in the text. The corpus provides a manually annotated Gold Standard generated by expert linguists. There are also annotation guidelines and consistency analysis to ensure the quality of the corpus.

The corpus has no explicit testing splits so we can consider that the train split has $29,682$ samples. The corpus has a \texttt{cc-by-sa-3.0} licence and is publically available \footnote{https://github.com/Vicomtech/NUBes-negation-uncertainty-biomedical-corpus}.

\subsubsection{PharmaCoNER} \label{data:pharmaconer}

The PharmaCoNER (Pharmacological Substances, Compounds and proteins and Named Entity Recognition) corpus \cite{gonzalez2019pharmaconer} is a random sample of the SPACCC corpus that has been annotated manually for two different tasks. 

\begin{itemize}
    \item NER: The entities are drugs and chemicals that appear in the text. Depending on their nature, they can be "normalizable" using SNOMED-CT, "non normalizable", "proteins" or "unclear".
    \item Concept indexing: Task to identify all the SNOMED-CT IDs corresponding to each of the entities from the first task.
\end{itemize}

The corpus provides a manually annotated Gold Standard generated by medicinal chemistry experts. There are also annotation guidelines and consistency analysis to ensure the quality of the corpus.

The corpus has a size of $500$, $250$ and $250$ train, development and test samples respectively. The test split comes from a bigger background unnanotated split of $2751$ unnanotated samples. The corpus has a \texttt{cc-by-4.0} licence and is publically available \footnote{https://zenodo.org/record/4270158}.

\subsubsection{SocialDisNER} \label{data:socialdisner}

SocialDisNER (Mining social media content for disease mentions) \cite{gasco2022socialdisner} is a collection of health-related tweets focused on disease mentions. Although it is not strictly a clinical corpus, its unique nature of informal language talking about health can be very useful to check the generalization of the models.

The corpus provides a manually annotated Gold Standard generated by experts. There are also annotation guidelines and consistency analysis to ensure the quality of the corpus.

The corpus has a size of $6000$, $2000$ and $2000$ train, development and test samples respectively. The corpus has a \texttt{cc-by-4.0} licence and is publically available \footnote{https://zenodo.org/record/6803567}.




\begin{table}[!ht]
\centering
\begin{tabular}{lcccc}
\hline
\textbf{corpus} & \multicolumn{1}{c}{\textbf{Train samples}} & \multicolumn{1}{l}{\textbf{Val samples}} & \textbf{Test Samples} & \textbf{Problem type} \\ \hline
BARR2        & 318    & 146  & 220  & NER \\
CANTEMIST    & 501    & 500  & 300  & NER \\
CARES        & 2250   & 0    & 996  & Classification \\
CWLC         & 9000   & 0    & 0    & NER \\
CodiEsp      & 500    & 250  & 250  & Classification \\
CT-EBM-SP    & 12600  & 4510 & 4550 & NER \\
DisTEMIST    & 750    & 0    & 250  & NER \\
eHealthKD    & 1500   & 50   & 150  & NER \\
IULA-SCRC    & 3194   & 0    & 0    & NER \\
IxaMed-GS    & 1875   & 1097 & 995  & NER \\
LivingNER    & 1000   & 500  & 485  & NER \\
MEDDOCAN     & 500    & 250  & 250  & NER \\
NUBes        & 29,682 & 0    & 0    & NER \\
PharmaCoNER  & 500    & 250  & 250  & NER \\
SocialDisNER & 6000   & 2000 & 2000 & NER \\ \hline
\end{tabular}
\caption{Summary of Spanish Clinical Corpora}
\label{tab:datasets}
\end{table}

\section{Models} \label{sec:models}

Despite some recent attempts on creating static word embeddings like Spanish Clinical Embeddings \cite{spanishclinicalembeddings}, the Word embeddings for the Spanish clinical language \cite{carolinachiu20226647060} and even contextual embeddings like the Spanish Clinical Flair \cite{rojas-etal-2022-clinical} we will not emphasize on those because it has been proven that transformer-based language models outperform them in most of the situations \cite{muennighoff2023mteb}.

However, given that there are not that many Spanish clinical language models, we will also list general models with good results in the Spanish language.

\subsection{BETO}

BETO \cite{beto} was the first Spanish language model with BERT architecture. The model was trained with the Spanish Unnanotated Corpora (SUC) \cite{suc} (4GB corpus) and has 110M parameters (the same as BERT). Although it was released in $2020$, the model is still useful as a baseline given its simplicity and stability.

The model has \texttt{cc-by-4.0} licence and is publically available \footnote{https://huggingface.co/dccuchile/bert-base-spanish-wwm-cased}.

\subsection{MarIA}

The next general Spanish language model is MarIA \cite{maria}. Released in $2021$, MarIA is based in the RoBERTa \cite{roberta} architecture. The base version of the model has 117M parameters and the large one has 355M, with the large version having the exact same architecture and size than the original RoBERTa-large. The corpus used to train this model is a web crawl performed by the Spanish National Library with a size of 570GB once clean.

Although both models have good results, this paper will focus on the large model given that the results are better and the model is still small enough to fit in a single GPU.

The model has \texttt{apache-2.0} licence and is publicly available \footnote{https://huggingface.co/PlanTL-GOB-ES/roberta-large-bne}.

\subsection{RigoBERTa}

RigoBERTa is a DeBERTa-based model released in $2022$ by the Instituto de Ingeniería del Conocimiento (IIC) \cite{serrano2022rigoberta}. The model has 134M parameters and was trained on a mix of various crawl sources (such as OSCAR \cite{oscar} and mC4 \cite{xue2021mt5}), Spanish news articles and the Spanish Unnanotated Corpora (SUC) \cite{suc}.

Later, in $2023$, the second version of the model has been released, RigoBERTa 2. This is the model that will be tested in the benchmark.

The model has all rights reserved and is not publically available, although it can be obtained through its autors.

\subsection{XLM-RoBERTa}

Before jumping into the specific clinical models, it is important to note that there are some very good performing multilingual models that we should take into account given the small amount of purely Spanish language models publicly available. Otherwise we could leave very good models out \cite{agerri2022lessons}.

XLM-RoBERTa \cite{xlmrcc100} is a multilingual version of RoBERTa trained in 2019. The model was trained with the CC100 corpus \cite{xlmrcc100}. This corpus has 2.5TB in size and has data of $100$ different languages.

There are several models trained with this architecture and corpus: base, large, xl and xxl sizes with 250M, 560M, 3.5B and 10.7B parameters respectively. For this paper we will stick with the large model given that it fits in a GPU and is stable enough given the sizes of the corpora used in the benchmark.

The model has \texttt{MIT} licence and is publically available \footnote{https://huggingface.co/xlm-roberta-large}.

\subsection{DeBERTaV3}

DeBERTaV3 \cite{he2021debertav3} is the latest DeBERTa \cite{he2021deberta} based model and it was released in 2021. Although DeBERTa is a family of English language models, DeBERTaV3 includes a multilingual model too, mDeBERTaV3. 

The model was trained with the CC100 corpus, the same as XLM-RoBERTa. The resulting multilingual model has 276M parameters which starts to be in the big side for a single GPU.

The model has \texttt{MIT} licence and is publically available \footnote{https://huggingface.co/microsoft/mdeberta-v3-base}.

\subsection{GPT family}

After the release of ChatGPT in late 2022 and GPT-4 in early 2023, this family of models brought a revolution both to the NLP research community and to the society in general \cite{reuters2023}. Although these are generative models, they shine for their ability to tackle any kind of problem in a zero or few shot learning approach.

These models are not open source and can only be accessed via the OpenAI API \footnote{https://platform.openai.com/docs/api-reference}. Therefore we can only make educated guesses about how they have been trained and with which corpora. There are also some ethical concerns regarding the closed nature of the models \cite{gpt4survey}, even bigger in this medical domain where a lot of medical institutions have strict security protocols that forbid them to upload personal data to the internet.

For instance, ChatGPT is an evolution of GPT-3 \cite{gpt3} fine-tuned with some kind of RLHF (Reinforcement Learning from Human Feedback) \cite{rlhf}. In the case of GPT-4 \cite{gpt4} we can assume that is an evolution of GPT-3 with a bigger corpora and more parameters.

There is evidence that GPT-4 has improved a lot the results in some clinical NLP benchmarks in English \cite{nori2023capabilities}, however it has not been tried in Spanish or in more encoder-focused tasks like Named Entity Recognition. In the Section \ref{sec:expgpt} we will try to use these models in some clinical corpora to check their viability in every aspect.

\subsection{Galén family}

The Galén \cite{galen} models are domain adaptations \cite{domainadapt} of preexisting Spanish capable language models like mBERT, BETO and XLM-RoBERTa published in 2021. This specific domain adaptation has been performed by further pretraining the above mentioned models with a clinical corpus extracted from the Galén Oncology Information System \cite{ribelles2010galen}.

All models are publically available under a \texttt{MIT} licence \footnote{https://github.com/guilopgar/ClinicalCodingTransformerES}, however they are not published in the Model Hub, only their weights are published which makes it harder to reuse them.

\subsection{bsc-bio-ehr-es}

bsc-bio-ehr-es \cite{carrino-etal-2022-pretrained} is a biomedical model based in the RoBERTa architecture published in 2022. The model has been trained from scratch (instead of a domain adaptation) using a large biomedical corpora and also a smaller corpus of electronic health records.

There are two models in these series, bsc-bio-es and bsc-bio-ehr-es where the first is trained only with the first corpora and the later is trained with all the gathered data. Both models have 125M parameters. In this paper we will only compare the bsc-bio-ehr-es model given that the authors report better results for this use case.

Most of the data sources for these models are not publically available so we cannot assert the claims of the authors.

The model has \texttt{apache-2.0} licence and is publically available \footnote{https://huggingface.co/PlanTL-GOB-ES/bsc-bio-ehr-es}.

\subsection{Other Generative Large Language Models}

Since the release of ChatGPT, a lot of new generative Large Language Models have been created. Although they are way out of scope for this study and most of them are in English, we will mention the most relevant ones that may be featured in future work. Also we carry out a preliminary study on chat models (OpenAI models) to check their feasibility in Section \ref{sec:expgpt}.

Most of the models have a focus in English language, but due to their size, they can all perform well in Spanish. Most notoriously we can find Llama \cite{touvron2023llama} and Llama2 \cite{touvron2023llama2} with some of their biggest versions being multilingual, Falcon \cite{falcon40b} with a 40B multilingual version, Vicuna \cite{vicuna}, BLOOM \cite{workshop2023bloom} and PaLM \cite{chowdhery2022palm}.

\section{Method} \label{sec:meth}

Formally, we will evaluate results using the F1 score for the NER corpora and F1 with micro average classification tasks. However we will also take into account the availability of the models and the computational cost of running them.

The main evaluation has been designed for the encoder models and they are fine-tuned using the chosen corpora. For the metaparameters optimization we designed a grid of parameters for all the models, described in the Appendix \ref{appendix:metaparameters}.

\subsection{Corpora}

Most of the clinical corpora in Spanish are either NER corpora or non-labeled corpora so there is a relatively low variety of corpora in order to make a benchmark. Having this into account, we selected the following corpora for the benchmark:

\begin{itemize}
    \item cantemist: Good example of well annotated NER corpus (\ref{data:cantemist}).
    \item caresA: Multi-label classification of the body area mentioned on the radiological report (\ref{data:cares}).
    \item caresC: Multi-label classification of the ICD-10 chapters that correspond to the radiological report (\ref{data:cares}).
    \item ctebmsp: (\ref{data:ctebmsp}).
    \item distemist: Good example of well annotated NER corpus  (\ref{data:distemist}).
    \item ehealth\_kd: Subtask A, NER with very different entities (\ref{data:ehealth}).
    \item livingner1: LivingNER task 1. Good example of well annotated NER corpus (\ref{data:livingner}).
    \item livingner3: LivingNER task 3. Multi-label classification corpus.(\ref{data:livingner}).
    \item meddocan: Anonymization corpus in clinical cases (\ref{data:meddocan}).
    \item nubes: Big NER corpus and with negation entities (\ref{data:nubes}).
    \item pharmaconer: Good example of well annotated NER corpus (\ref{data:pharmaconer}).
    \item socialdisner: Informal language talking about health (\ref{data:socialdisner}).
\end{itemize}


\subsection{Models}

Due to the limited time and computational budget, not every available model has been tested in the benchmark. Next we will list the chosen models:

\begin{itemize}
    \item BETO: We use BETO as a baseline for Spanish language models.
    \item BETO\_Galén: A domain adaptation of BETO, it should be better for these experiments.
    \item bsc-bio-ehr-es: Smaller and trained only with relevant data.
    \item MarIA: The best open-sourced Spanish only encoder language model.
    \item mDeBERTaV3: Second-best performing multilingual model.
    \item RigoBERTa 2: The best Spanish only encoder language model.
    \item XLM-R\_Galén: A domain adaptation of the best model for Spanish.
    \item XLM-RoBERTa-Large: The best results for Spanish language.
\end{itemize}

We will also show some experiments with GPT-3 and GPT-4 that we did not include in the final benchmark due to some inherent limitations to generative language models and the limited OpenAI API functionality.

\subsection{Public benchmark}

For the sake openness, all the final fine-tuned models, as well as all the corpora used and the code for the benchmark will be publically available \footnote{https://github.com/iiconocimiento/survey-spanish-clinical-language-models}. Note that neither the RigoBERTa 2 fine-tuned models nor the base model can be publically available due to license constrains.

Reuploading the corpora and gathering them all in the Hugging Face Hub allows us to use the Hugging Face Leaderboards \footnote{https://huggingface.co/spaces/autoevaluate/leaderboards} to create the first benchmark for Spanish Clinical Language Models.

All the information about the resources will be displayed in Appendix \ref{appendix:open}.

\section{Evaluation and results} \label{sec:bench}

In Table \ref{tab:results} we can see the results of the benchmark. RigoBERTa 2 obtains the best results by far, winning in $6$ out of the $12$ corpora. Overall it was also the most stable model, consistently obtaining good results, not like mDeBERTaV3 that, despite having very good results, obtained an exceptionally bad result in the livingner3 corpus. XML-RoBERTa-Large obtains also very good results and is the best open-sourced model available.

Paradoxically, the more specific the model is, the worse that its results are. At a first glance it might look like the bigger model always wins, however we must remember that both XLM-R\_Galén and BETO\_Galén are domain adaptations of bigger models. Specifically XLM-R\_Galén uses as base model XML-RoBERTa-Large, the best performing model in the benchmark. Consequently this means that it is not always worth to adapt a model to a certain domain if the corpus for the adaptation is not big or good enough.

bsc-bio-ehr-es, the only model trained from scratch in this benchmark obtains relatively good results given its size and the relatively small amount of data used during the training process. However, this model does not obtain any improvement against the general Spanish language models or the multilingual ones.

\begin{table}[!ht]
\centering
\begin{tabular}{l|cccccc|cc}
\hline
             & \multicolumn{6}{c|}{\textit{Spanish Only models}}                                                                     & \multicolumn{2}{c}{\multirow{2}{*}{\textit{Multilingual}}} \\
             & \multicolumn{3}{c:}{\textit{Clinical models}}                          & \multicolumn{3}{c|}{\textit{General Models}} & \multicolumn{2}{c}{}        \\ \hline
Corpus      & XLM-RG       & BETO\_G     & \multicolumn{1}{l:}{bsc-ehr}              & BETO             & MarIA                    & RigoB2               & mDeB3                  & XLM-RL                       \\ \hline
cantemist    & 0.898        & 0.802       & \multicolumn{1}{l:}{0.864}                & 0.898            & 0.902                    & {\ul 0.903}          & 0.890                  & {\ul \textbf{0.904}}         \\
caresA       & 0.989        & 0.977       & \multicolumn{1}{l:}{0.991}                & 0.992            & 0.992                    & {\ul \textbf{0.997}} & 0.993                  & {\ul 0.994}                  \\
caresC       & 0.823        & 0.551       & \multicolumn{1}{l:}{{\ul \textbf{0.862}}} & 0.835            & 0.840                    & 0.854                & 0.756                  & {\ul 0.847}                  \\
ctebmsp      & 0.881        & 0.726       & \multicolumn{1}{l:}{0.876}                & 0.880            & 0.877                    & {\ul \textbf{0.907}} & 0.902                  & {\ul 0.906}                  \\
distemist    & 0.759        & 0.346       & \multicolumn{1}{l:}{0.759}                & 0.801            & 0.793                    & {\ul \textbf{0.832}} & 0.808                  & {\ul 0.817}                  \\
ehealth\_kd  & 0.830        & 0.658       & \multicolumn{1}{l:}{0.836}                & 0.843            & 0.836                    & {\ul 0.865}          & 0.844                  & {\ul \textbf{0.871}}         \\
livingner1   & 0.907        & 0.646       & \multicolumn{1}{l:}{0.938}                & 0.938            & 0.939                    & {\ul 0.951}          & {\ul \textbf{0.953}}   & 0.949                        \\
livingner3   & 0.500        & 0.000       & \multicolumn{1}{l:}{0.604}                & 0.626            & {\ul \textbf{0.644}}     & 0.621                & 0.153                  & {\ul 0.606}                  \\
meddocan     & 0.947        & 0.682       & \multicolumn{1}{l:}{0.967}                & 0.957            & 0.977                    & {\ul \textbf{0.979}} & 0.974                  & {\ul 0.978}                        \\
nubes        & 0.908        & 0.762       & \multicolumn{1}{l:}{0.903}                & 0.908            & 0.911                    & {\ul 0.915}          & 0.919                  & {\ul \textbf{0.920}}         \\
pharmaconer  & 0.915        & 0.708       & \multicolumn{1}{l:}{0.904}                & 0.908            & 0.914                    & {\ul \textbf{0.927}} & 0.922                  & {\ul 0.924}                  \\
socialdisner & 0.919        & 0.777       & \multicolumn{1}{l:}{0.921}                & 0.915            & 0.920                    & {\ul \textbf{0.943}} & 0.935                  & {\ul 0.941}                  \\ \hline
Average      & 0.853        & 0.621       & \multicolumn{1}{l:}{0.869}                & 0.873            & 0.877                    & {\ul \textbf{0.890}} & 0.833                  & {\ul 0.887}                  \\
Wins group   & 0            & 0           & \multicolumn{1}{l:}{1}                    & 0                & 1                        & {\ul \textbf{10}}    & 1                      & {\ul 11}                     \\
Wins total   & 0            & 0           & \multicolumn{1}{l:}{1}                    & 0                & 1                        & {\ul \textbf{6}}                    & 1                      & {\ul 3}                   \\ \hline
\end{tabular}
\caption{Results with models grouped into Spanish Only and Multilingual models. Spanish Only models are splitted into Clinical models and General models. Best results for every group are \uline{underlined} and best results overall are in \textbf{bold}. We also report the sum of the scores for every model, wins in own group and wins overall. The metric is micro F1 for classification models and F1 for NER models. XLM-RF = XLM-R\_Galén, BETO\_G = BETO\_Galén, bsc-ehr = bsc-bio-ehr-es, RigoB2 = RigoBERTa 2, mDeB3 = mDeBERTaV3, XLM-RL = XLM-RoBERTa-Large}
\label{tab:results}
\end{table}

\begin{figure}[!ht]
\centering
     \includegraphics[width=\textwidth]{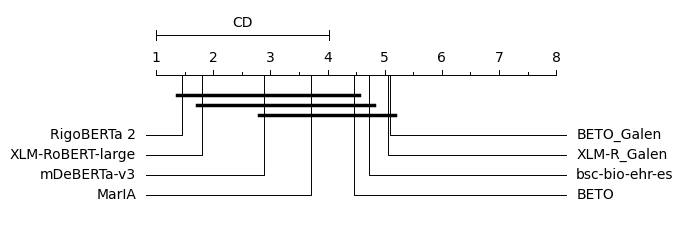}
      \caption{Nemenji plot of the results}
       \label{fig:nemenji}
\end{figure}

In the Figure \ref{fig:nemenji} we can see a Nemenji plot with a classification of the models given the results of the benchmark. We can see that, although the first models are within the critical distance, both RigoBERTa 2 and XLM-RoBERTa-Large are statistically better than all the domain adaptations.


\subsection{GPT family}\label{sec:expgpt}

We made two attempts to use this technology, first with the NER corpus CANTEMIST and then with the multi-label classification corpus CaresC. Generative models are not prepared for any of these tasks so we had to tune the prompts in order to achieve what we wanted. As we will see with the results of these experiments, it is not easy to use these models properly, but it is worth giving them a chance \cite{bang2023multitask}.

\subsubsection{Named Entity Recognition}

First, in the NER corpus we observed that the models (both GPT-3 and GPT-4) are capable of finding some entities in the text. However there is no way to obtain a comparable score if those entities do not have their respective positions also labeled in the text. It is technically possible to perform token classification tasks such as NER with generative language models: we just need to add a classification head over each output embedding. However, the OpenAI API does not allow to do it straightforwardly and transforms the task into a text-to-text problem which turned out to be inefficient and error prone.

While the models more or less understood our queries, there were a lot of formatting errors and most of the times every entity was positioned wrong. In Appendix \ref{appendix:gptner} a more detailed description of both the prompts and parameters used can be found.

After a lot of attempts without a good solution we exhausted our budged for this experiment and we decided to stop experimenting and not try any fine-tuning or few shot evaluation due to the high risk of a bad result and the high cost of the platform. This budget, similarly with the one showed in Table \ref{tab:resultsgpt}, was higher that what would have costed a GPU machine to fine-tune an encoder model.

We also tried the edits endpoint of the API to insert the NER entity after its appearance, but it did not work well either.

\subsubsection{Classification}

Next we tried with a multi-label classification corpus CaresC. Although it is relatively simple to correctly prompt these models in order to perform binary or multiclass classification, more precise prompting had to be done to achieve consistent and good results. All the details of these process can be found in the Appendix \ref{appendix:gptclassification}.

Once we obtained good enough prompts, we evaluated the performance in a few-shot approach. Also a fine-tunning was attempted, taking into account that the fine-tuning API does not support multi-label classification, so the training process was a general text generation fine-tuning. In the Table \ref{tab:resultsgpt} we can see a detailed summary of the results and costs.

\begin{table}[!ht]
\centering
\begin{tabular}{llc}
\hline
\textbf{model}                  & \multicolumn{1}{c}{\textit{\textbf{F1}}} & \multicolumn{1}{l}{\textbf{Cost (\$)}} \\ \hline
text-davinci-003 (few-shot)     & 0.268                                    & 11                                     \\
gpt-4 (few-shot)         & 0.386                                    & 46                                     \\
bsc-bio-ehr-es
& 0.862                                    & 20                                     \\
text-ada-001(fine-tune)         & 0.506                                    & 5                                      \\
text-davinci-003 (fine-tune)    & 0.621                                    & 166                                    \\ \hline
\end{tabular}
\caption{Table showing the micro F1 obtained by the models and their cost in the CaresC corpus. The cost for the bsc-bio-ehr-es was calculated taking into account the duration of the training process and the cost of a p4d.24xlarge in AWS. Note that gpt-4 is a chat-only model, not a completion model.}
\label{tab:resultsgpt}
\end{table}

Surprisingly, we can see that when comparing GPT models against a far simpler encoder-only model such as bsc-bio-ehr-es, the latter obtains the best results by far. Within GPT models, there is a clear increase of quality when a fine-tuning process is performed to the base models, up to a point that the few shot models appear as a poor-quality but expensive alternative. Even when using the most expensive GPT approach (fine-tune the Davinci model), we obtain a significantly poor result in comparison to a classic encoder fine-tuning. The only acceptable exception to all these bad results is the fine-tuning of the Ada model, which is $4$ times cheaper than bsc-bio-ehr-es; however the result in terms of quality is considerably worse.

There is also the concern that all the data used in the OpenAI API can be privately used by OpenAI to improve their models, which can make these solutions unusable for the clinical domain given the European data protection laws.

All in all, these results show that this Large Generative Language Models revolution does not apply to all fields, problem types and languages. There is also more evidence of this in more thorough studies like the one by Chen et al. \cite{chen2023large}. Wang et al. \cite{wang2023clinicalgpt} (similarly to Luo et al. \cite{Luo2022} with BioGPT, instead trained a ClinicalGPT in order to greatly improve the results in clinical domain tasks). However it must be noted that all these efforts generated English language models so they can be seen as future work for the Spanish language.

\section{Conclusion}

Through this study we have observed that there is a fair amount of corpora and models for the application of NLP to clinical problems in Spanish. However a lot of that corpora is not publically available or does not have a proper Gold Standard so models can easily learn from it and be evaluated.

We benchmarked the most promising models for the clinical domain in Spanish in a variety of corpora to check their real performance, and we found out that none of them is good enough to beat a big general multilingual model or a closed-source commercial one. This holds true even if the clinical models are domain adaptations of the best model open-source model. This means that there is a lot of work to be done for the Spanish language in terms of publicly available language models, and that we have room to improve.

Following those experiments, our main contribution was the release of a standardized benchmark to test future Spanish language models for medical applications and the gathering of a series of models and corpora in an easy to access platform.

We also tried the revolutionary generative models and proved that they cannot be used for all use cases: at this time there is no such thing as an Artificial General Intelligence and there are more narrow and affordable solutions that perform better for certain applications.

It is clear then, that we need better Spanish language models, both general and domain specific. We also need a big corpora of Spanish clinical texts so these models can be trained. We are aware that this is a big claim, but we just cannot relegate ourselves to using multilingual models in this crucial domain where people's lives are at stake.

\section*{Acknowledgments}
This work was supported in part by the Instituto de Ingeniería del Conocimiento and R\&D\&i ACCESS2MEET (PID2020-116527RB-I0) project supported by MCIN AEI/10.13039/501100011033/. Additionally, this work has been supported by "Intelligent and interactive home care system for the mitigation of the COVID-19 pandemic" PRTR-REACT UE project. 

\bibliographystyle{unsrt}  
\bibliography{references}  

\appendix

\section{Metaparameters}\label{appendix:metaparameters}

All the models in the evaluation have been trained with all the configurations from the Table \ref{tab:metaparameters}. Each one of the $7$ models has been trained with every combination of the parameters for all the $12$ corpora used for the evaluation, in total $3024$ models were trained.

\begin{table}[!ht]
\centering
\begin{tabular}{ll}
\hline
\textbf{Metaparameter}  & \textbf{Configurations}      \\ \hline
Batch Size              & $\{16, 32, 64\}$             \\
Classifier Dropout      & $\{0, 0.1, 0.2\}$            \\
Learning Rate           & $\{1e-5, 2e-5, 3e-5, 4e-5\}$ \\
Warmup Ratio            & $0$                          \\
Warmup Steps            & $0$                          \\
Weight Decay            & $0$                          \\
Optimizer               & \textit{AdamW}               \\
Epochs                  & $10$                         \\
Early Stopping Patience & $3$                          \\ \hline
\end{tabular}
\caption{Parameter grid used for the experiments}
\label{tab:metaparameters}
\end{table}

\section{Public Resources}\label{appendix:open}

\subsection{Corpora} \label{appendix:open:datasets}

Some corpora did not have a labeled test split so some of the splits used do not coincide with the original splits in the corpora. In the following list we can see a list of the used corpora and their corresponding Huggingface Hub repository. Note that the name of the repository is the one that was used for the fine-tuned models.

\begin{itemize}
    \item cantemist: \texttt{PlanTL-GOB-ES/cantemist-ner}
    \item caresA: \texttt{chizhikchi/CARES}
    \item caresC: \texttt{chizhikchi/CARES}
    \item ctebmsp: \texttt{lcampillos/ctebmsp}
    \item distemist: \texttt{bigbio/distemist}
    \item ehealth\_kd: \texttt{ehealth\_kd}
    \item livingner1: \texttt{IIC/livingner1}
    \item livingner3: \texttt{IIC/livingner3}
    \item meddocan: \texttt{bigbio/meddocan}
    \item nubes: \texttt{plncmm/nubes}
    \item pharmaconer: \texttt{PlanTL-GOB-ES/pharmaconer}
    \item socialdisner: \texttt{IIC/socialdisner}
\end{itemize}

\subsection{Models}

First of all, we uploaded both BETO Galén \footnote{https://huggingface.co/IIC/BETO\_Galen} and XLM-R Galén \footnote{https://huggingface.co/IIC/XLM-R\_Galen} to the Model Hub because they were only available through they raw weights in GitHub.

Every fine-tuned model follows the structure \texttt{{model-name}-{corpus}}, so for example, the model id for MarIA and the CANTEMIST corpus would be \texttt{IIC/roberta-large-bne-cantemist} \footnote{https://huggingface.co/IIC/roberta-large-bne-cantemist}. So combining the models from the list below and the corpus names from the list in Appendix \ref{appendix:open:datasets} lead to their corresponding Huggingface Hub address.

\begin{itemize}
    \item BETO: \texttt{IIC/bert-base-spanish-wwm-cased-\{corpus\}}
    \item BETO\_Galen: \texttt{IIC/BETO\_Galen-\{corpus\}}
    \item MarIA: \texttt{IIC/roberta-large-bne-\{corpus\}}
    \item XLM-RoBERTa-Large: \texttt{IIC/xlm-roberta-large-\{corpus\}}
    \item XLM-R\_Galen: \texttt{IIC/XLM\_R\_Galen-\{corpus\}}
    \item bsc-bio-ehr-es: \texttt{IIC/bsc-bio-ehr-es-\{corpus\}}
    \item mDeBERTaV3: \texttt{IIC/mdeberta-v3-base-\{corpus\}}
\end{itemize}

\subsection{Code Repository}

All the code used for the benchmark is publically available in GitHub \footnote{https://github.com/iiconocimiento/survey-spanish-clinical-language-models}. Note that the code uses some data and models from their raw origin because the benchmark was done before uploading every model and corpus to the model hub.

\section{GPT-3 experiments}\label{appendix:gpt}

Here we will explain in more detail the experiments with the OpenAI API.

\subsection{NER experiments}\label{appendix:gptner}

In the following snippet we can see an example prompt introduced to the chat model where \texttt{example1} and \texttt{example2} are few shot samples and \texttt{text} is the sample that has to be predicted:

\noindent\fbox{
    \parbox{\textwidth}{
Recibes un texto y tu tarea es reconocer y extraer las entidades específicas del tipo
`ENFERMEDAD`. \\ \\ Tu formato de salida es el siguiente: [[inicio, fin, 'ENFERMEDAD'], [inicio, fin, 'ENFERMEDAD'], 

..., [inicio, fin, 'ENFERMEDAD'] 

Donde inicio y fin es la posición del texto que ocupa cada enfermedad.

Ejemplos:\\ \\ \{example1\}\\ \\ \\ \{example2\}\\ \\ Texto: \{text\}\\ Enfermedades:

    }
}

When using the chat API we also used the \texttt{system} prompt to bias the model towards our use case:

\noindent\fbox{
    \parbox{\textwidth}{
Eres un sistema de Named Entity Recognition (NER) extremadamente inteligente y preciso
en el dominio biomédico.
    }
}

A simplified version of the above prompt was also tried, however the results could not be compared to the Gold Standard because of the lack of the entity positions:

\noindent\fbox{
    \parbox{\textwidth}{
Identifica y las enfermedades que aparecen en los siguientes textos. Puede haber
enfermedades repetidas. \\ Ejemplos:\\ \\ \{example1\}\\ \\ \\ \{example2\}\\ \\ Texto:
\{text\} \\ Enfermedades:

    }
}

Some parameter combinations were tried, however the best performing one was: \texttt{temperature=0}, \texttt{top\_p=0.8}, \texttt{presence\_penalty=-2} and \texttt{frequency\_penalty=0}. Logit biases were also introduced, but they made the model reach local minima very fast resulting in nonsense predictions.

\subsection{Classification experiments}\label{appendix:gptclassification}

Here we followed 2 different approaches, few-shot evaluation of the test data and fine-tuning.

Like for the NER corpus, for the few-shot approach we tried with some prompts with examples and \texttt{system} prompts:

\noindent\fbox{
    \parbox{\textwidth}{
Recibes un texto y tu tarea es elegir todos los capítulos del standard ICD-10 al que
pertenece (números del 1 al 21).

Ejemplos:\\ \\ \{example0\}\\ \\ \{example1\} \\ \\ \{example2\} \\ \\ \\ Entrada: \{text\} -> \\ \\ \#\#\# \\ \\
    }
}

Where an \texttt{example} could be the text and the chapters in a Python list syntax:

\noindent\fbox{
    \parbox{\textwidth}{
\{text\} -> [11, 14]
    }
}

In this case, the \texttt{system} prompt was:

\noindent\fbox{
    \parbox{\textwidth}{
Eres un sistema experto en códigos ICD-10 extremadamente inteligente y preciso en el
dominio biomédico.
    }
}

Some parameter combinations were tried, however the best performing one was: \texttt{temperature=0}, \texttt{top\_p=0.8}, \texttt{presence\_penalty=0} and \texttt{frequency\_penalty=0}. Logit biases were also introduced, the model is sensitive to the bias but we achieved some extra consistency with a bias of $1$ for every possible class (the tokens for the numbers of the chapters) and $0.2$ for the comma, opening and closing bracket tokens.

For the fine-tuning approach we followed the steps and good practices explained in the OpenAI API documentation. We used the same parameters and the same prompts used for the few-shot evaluation.

\end{document}